%% file: main.tex
\documentclass[10pt]{article}

\usepackage[utf8]{inputenc}
\usepackage[T1]{fontenc}
\usepackage[english]{babel}

\usepackage[left=2cm,right=2cm,top=2.25cm,bottom=2.25cm,headheight=12pt,letterpaper]{geometry}

\usepackage{ifthen,calc}
\usepackage{microtype}

\usepackage[fleqn]{amsmath}
\usepackage{amsfonts,amssymb}
\usepackage{newtxtext}
\usepackage{newtxmath}
\usepackage[scaled=0.92]{helvet}
\usepackage{courier}

\usepackage{graphicx,xcolor}
\usepackage{booktabs}
\usepackage{multirow}
\usepackage{array}
\usepackage{subcaption}

\usepackage{authblk}
\usepackage[labelfont={bf,sf},labelsep=period,justification=raggedright]{caption}
\usepackage[colorlinks=true,allcolors=blue]{hyperref}
\usepackage[superscript,nomove]{cite}

\usepackage{fancyhdr}
\usepackage{lastpage}
\usepackage[explicit]{titlesec}
\titleformat{\section}
  {\color{black}\large\sffamily\bfseries}
  {\thesection}{0.5em}{#1}
\titleformat{name=\section,numberless}
  {\color{black}\large\sffamily\bfseries}
  {}{0em}{#1}
\titleformat{\subsection}
  {\sffamily\bfseries}
  {\thesubsection}{0.5em}{#1}
\titleformat{\subsubsection}
  {\sffamily\small\bfseries\itshape}
  {\thesubsubsection}{0.5em}{#1}
\titleformat{\paragraph}[runin]
  {\sffamily\small\bfseries}
  {}{0em}{#1}

\titlespacing*{\section}{0pt}{3ex}{5pt}
\titlespacing*{\subsection}{0pt}{2.5ex}{0pt}
\titlespacing*{\subsubsection}{0pt}{2ex}{0pt}
\titlespacing*{\paragraph}{0pt}{1.5ex}{10pt}

\makeatletter
\renewcommand{\@maketitle}{%
\thispagestyle{empty}

\begin{center}
{\Large\sffamily\bfseries \@title\par}
\vspace{1em}
{\normalsize\sffamily \@author\par}
\vspace{1.5em}
\end{center}

\noindent{\bfseries\sffamily Abstract}

\vspace{0.5em}

\small
\@abstract

\vspace{1.5em}
}
\newcommand{\abstracttext}[1]{\gdef\@abstract{#1}}
\makeatother

\title{Robust-ComBat: Mitigating Outlier Effects in Diffusion MRI Data Harmonization}

\author[1,*]{Yoan David}
\author[1,2]{Pierre-Marc Jodoin}
\author[+]{Alzheimer’s Disease Neuroimaging Initiative}
\author[ ]{The TRACK-TBI Investigators}

\affil[1]{VitaLab, Dep of Computer Science, University of Sherbrooke, Sherbrooke, QC, J1K 2R1, Canada}
\affil[2]{Imeka Solutions Inc., Sherbrooke, J1H 4A7, QC, Canada}

\affil[*]{yoan.david@usherbrooke.ca}

\affil[+]{
A comprehensive list of authors and affiliations of this consortium appear at the end of the paper. ida@loni.usc.edu.
}

\abstracttext{
Harmonization methods such as ComBat and its variants are widely used to mitigate diffusion MRI (dMRI) site-specific biases. However, ComBat assumes that subject distributions exhibit a Gaussian profile. In practice, patients with neurological disorders often present diffusion metrics that deviate markedly from those of healthy controls, introducing pathological outliers that distort site-effect estimation. This problem is particularly challenging in clinical practice as most patients undergoing brain imaging have an underlying and yet undiagnosed condition, making it difficult to exclude them from harmonization cohorts, as their scans were precisely prescribed to establish a diagnosis.

In this paper, we show that harmonizing data to a normative reference population with ComBat while including pathological cases induces significant distortions. Across 7 neurological conditions, we evaluated 10 outlier rejection methods with 4 ComBat variants over a wide range of scenarios, revealing that many filtering strategies fail in the presence of pathology. In contrast, a simple MLP provides robust outlier compensation enabling reliable harmonization while preserving disease-related signal.

Experiments on both control and real multi-site cohorts, comprising up to $80\%$ of subjects with neurological disorders, demonstrate that \textit{Robust-ComBat} consistently outperforms conventional statistical baselines with lower harmonization error across all ComBat variants.
}

\begin{document}
\maketitle

\section*{Introduction}
\label{sec:introduction}

Diffusion MRI (dMRI) provides sensitive markers of white matter microstructure and has been extensively applied in research on neurological and psychiatric disorders~\cite{Assaf2008,Jones2013}. However, data collected across sites are strongly affected by scanner differences, acquisition protocols, and demographic heterogeneity~\cite{Mirzaalian2016,Fortin2017}. These inter-site biases can obscure or even overshadow the biological effects of interest, rendering naive pooling of multi-site data problematic. As a result, harmonization has emerged as a critical step in modern neuroimaging pipelines.

Among the available harmonization approaches, ComBat~\cite{Fortin2017,Fortin2018} has gained prominence due to its simplicity, interpretability, and effectiveness. Originally designed for genomic data~\cite{Johnson2007}, ComBat applies an empirical Bayes model to remove additive and multiplicative site effects while preserving covariate-related biological variation. Its success in neuroimaging has led to several extensions like CovBat~\cite{Chen2021CovBat}, ComBat-GAM~\cite{Pomponio2020}, ComBat-GMM~\cite{Horng2022GMMComBat}, Pairwise-ComBat~\cite{Jodoin2025}, Clinical-ComBat~\cite{Girard2025}, and many others. Despite their methodological diversity, they share a critical assumption: that site-level data are drawn from clean and representative distributions.

\begin{figure}
    \centering
    \includegraphics[width=0.6\linewidth]{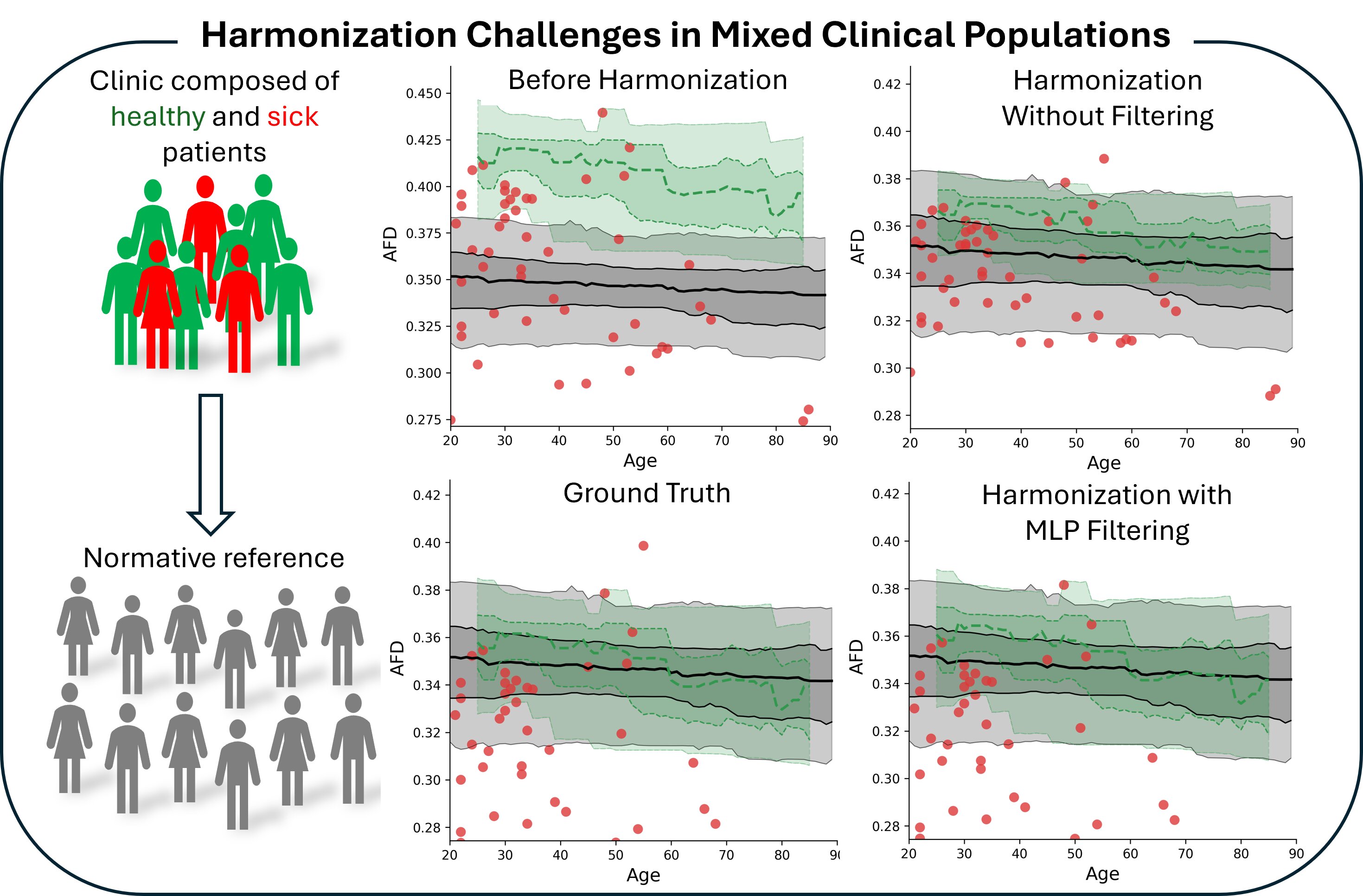}
    \caption{Illustration of the effect of mixing healthy controls (HC, green shading) and pathological subjects (TBI, red dots) from a given site on ComBat harmonization of AFD in the left IFOF. Prior to harmonization, the HC distribution from this site shows a marked deviation from the normative reference (gray). When HC and TBI subjects are blindly included in the harmonization process using a site composed of 50\% TBI patients (top right), both populations are artificially compressed toward the normative distribution. In contrast, when pathological outliers are properly filtered out from the estimation of site effects, harmonization is near optimal: HC distributions align with the normative reference, while TBI subjects remain consistently shifted below it.}
    \label{fig:clinicalcontext}
\end{figure}

However, this assumption is systematically violated in day-to-day clinical practice. Multi-site datasets arising from clinical practice contain a large proportion of pathological outliers, i.e., patients with neurological disorders whose diffusion profiles deviate substantially from those of healthy controls.  For instance, in certain white matter bundles, the mean free water (FW) in Alzheimer's disease (AD)~\cite{Ji2017} and the mean axial fiber diffusivity (AFD) in traumatic brain injury (TBI)~\cite{Wallace2020HBM} deviate by more than two standard deviations from healthy controls.  As shown by Jodoin et al.~\cite{Jodoin2025}, when such cases are included in site effect estimation, the harmonization process can be compromised.  This situation is illustrated in Figure~\ref{fig:clinicalcontext}.   Moreover, due to the cost, logistical burden, and scanner time required for MRI acquisition, clinical sites are typically unwilling to acquire additional scans from healthy controls solely for the purpose of harmonization.

While quality control pipelines~\cite{Oguz2014DTIPrep,Bastiani2019QC} exist to filter motion artifacts or detect gross acquisition failures, they are ill-suited for addressing the clinical reality of pathological outliers. Some recent work has attempted to extend harmonization frameworks to better handle outliers. For instance, BAMBOO~\cite{Smits2025BAMBOO} introduced an interquartile range filter to identify abnormal values, and related approaches in genomics such as ComBat-Seq~\cite{Zhang2020ComBatSeq} have sought to adapt to different data distributions. Yet these methods remain limited in scope as they rely on simple statistical heuristics, lack sensitivity to the complex patterns expressed by patients with neurological disorders, and are not directly applicable to dMRI.

In this paper, we propose {\em Robust-ComBat}, a harmonization framework tailored to the reality of routine clinical practice.  We demonstrate that applying ComBat harmonization to data that include pathological cases, when aligned to a normative reference, can introduce marked distortions in the harmonized measures. To investigate this issue in depth, we performed an extensive empirical study combining four variants of ComBat with ten outlier rejection methods, evaluated across a wide range of scenarios involving subjects with six neurological disorders. This analysis reveals that outlier filtering strategies differ substantially in their robustness to pathology, with several commonly used approaches failing to prevent harmonization bias. In contrast, a simple multilayer perceptron (MLP) consistently shows strong performance in compensating for pathological outliers.

\section*{Previous works}

\subsection*{Harmonization with ComBat and its Variants}

The ComBat family of methods has become the standard for harmonizing dMRI metrics such as fractional anisotropy (FA), the mean diffusivity (MD), the FW, and many others. Initially developed for genomics and later adapted to neuroimaging, ComBat and its variants aim to remove site effects while preserving biological variability. 

\textbf{ComBat} \cite{Fortin2017} is the foundation of several harmonization frameworks. It models and removes scanner effects using additive and multiplicative terms while preserving biological covariates such as age or sex. By estimating these site effects through an empirical Bayesian model, it stabilizes parameter estimation when site sizes are small. However, ComBat assumes that each site's data follow a roughly Gaussian distribution and represent a clean, homogeneous population, an assumption rarely met in clinical datasets containing outliers or pathological subjects.

\textbf{Pairwise-ComBat} \cite{Jodoin2025} follows the same statistical formulation as the original ComBat model. However, instead of aligning all sites toward a common pooled distribution, each moving site is independently projected onto a predefined normative reference population by removing its additive and multiplicative batch effects and replacing them with those of the reference site.

\textbf{CovBat} \cite{Chen2021CovBat} is one of the most commonly-cited extensions of ComBat.  It reformulates the mathematical foundation of ComBat to include covariance structures across sites. This ensures that inter-feature correlations (e.g., between brain regions) are consistent after harmonization, which is useful for multivariate and connectomic analyses. 

\textbf{ComBat-GAM} \cite{Pomponio2020} introduces nonlinear modeling of covariates, especially age, using generalized additive models (GAMs). This allows it to better capture non-linear developmental or aging trends in the data. 

\textbf{ComBat-GMM}~\cite{Horng2022GMMComBat} extends ComBat to more heterogeneous datasets by modeling each site as a Gaussian mixture model (GMM) rather than a single Gaussian distribution. This approach captures bimodal or multi-modal site-specific distributions that may arise from unobserved batch effects. However, it does not explicitly identify or handle outliers and assumes that all subjects belong to one of the modeled Gaussian components. Since outliers typically deviate from Gaussian behavior, even a modified version of ComBat-GMM would be unlikely to effectively address them.

\textbf{Clinical-ComBat}~\cite{Girard2025} is designed to handle real-world clinical scenarios by harmonizing each site independently to a normative reference population, allowing flexible integration of new data and clinics. It employs a polynomial data model and adaptive variance priors tailored to heterogeneous cohorts. 

\textit{Common Limitation.}
All ComBat variants share a key limitation: they assume that each site provides clean, Gaussian-like data representative of the same underlying population. In everyday clinical practice, however, dMRI datasets are dominated by pathological cases, i.e., patients with conditions such as AD, TBI, or other neurological disorders. In most clinical settings, the health status of these patients is often unknown, as diagnoses are typically established a posteriori by external specialists. This uncertainty makes it difficult to identify which subjects are truly pathological, thereby violating model assumptions and biasing the estimation of site effects, which can lead to over- or under-correction and loss of clinically relevant information.

\subsection*{Outlier-Aware Harmonization}

Few studies have explicitly addressed outlier handling within multi-site harmonization frameworks. Existing approaches typically rely on generic filtering strategies that fail to account for the complexity of clinical deviations or the multivariate nature of imaging data.

\textbf{BAMBOO} \cite{Smits2025BAMBOO} is a batch correction method for proteomic PEA data. It models three batch-effect types (protein-specific, sample-specific, plate-wide) and corrects them using bridging controls across plates, robust regression for plate-wide bias, and per-protein median adjustments. It also applies a univariate interquartile range (IQR) filter to remove extreme values in the bridging controls and flags proteins near the limit of detection. 

Unfortunately, BAMBOO does not translate to multi-site medical imaging. dMRI features are continuous, high dimensional, and strongly multivariate across fiber tracts and metrics, and pathological subjects often deviate in subtle, correlated ways that a univariate IQR screen can hardly capture. As will be shown, IQR filtering can modestly improve results, but it does not yield substantial gains when pathology induces small-to-moderate shifts across many correlated dimensions. Imaging pipelines also lack plate-like bridging controls, further limiting direct applicability.

\textbf{ComBat-Seq}~\cite{Zhang2020ComBatSeq} extends the ComBat framework to count-based data for RNA-seq gene expression profiles. Its statistical model replaces the Gaussian assumption of traditional ComBat with a negative binomial distribution, which better captures the discrete and overdispersed nature of sequencing counts. 

However, these assumptions are specific to RNA-seq data and do not hold in medical imaging, where features such as FA, MD, or FW are continuous, spatially structured, and approximately Gaussian. As a result, the probabilistic foundation of ComBat-Seq is ill-suited for imaging data, making the method irrelevant to the challenges of harmonization and outlier management in neuroimaging.

\section*{Context}

\subsection*{Clinical application}

In clinical settings, data harmonization is constrained by practical realities: MRI scan time is expensive, waiting lists are long, and acquiring large cohorts of healthy volunteers solely for harmonization is unrealistic. Clinical sites operate fixed protocols dictated by routine care, requiring software solutions to adapt rather than the opposite. Although harmonization should in principle rely on 16 to 32 scans from healthy controls per site\cite{Jodoin2025}, harmonization data are typically drawn from patients undergoing diagnostic evaluation, with an unknown and potentially high prevalence of pathology.  If not handled with care, these pathological data can can act as outliers and bias harmonization.

\subsection*{Outliers in Multi-site Diffusion MRI}

As described in \cite{Han2023Frontiers}, two main sources of outliers are typically distinguished: acquisition artefacts and pathology-related abnormalities. Artefactual outliers originate from acquisition and/or data processing issues \cite{VanDijk2012,Reuter2015,Ma2022OutlierMRI}. In contrast, pathology-driven outliers reflect genuine microstructural deviations associated with disease, such as neuronal loss, demyelination, or axonal degeneration. 
In this study, we focus specifically on the pathology-driven outliers. 

\begin{figure}[!t]
\centerline{\includegraphics[width=0.99\columnwidth]{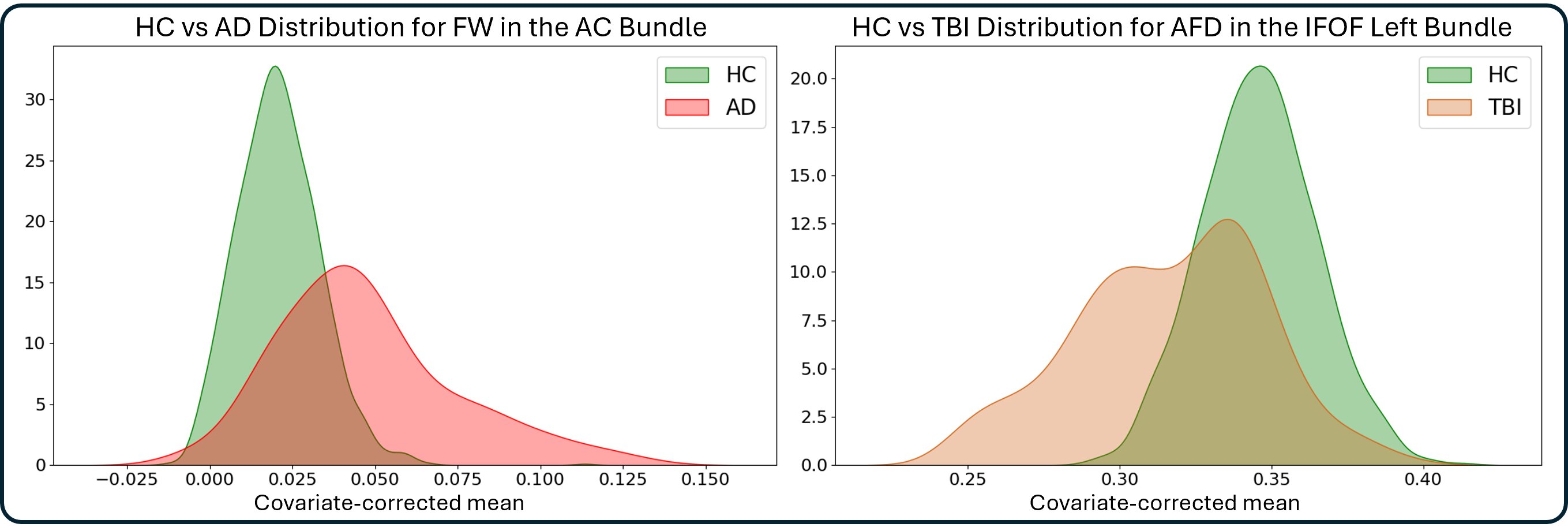}}
\caption{Illustration of pathology-driven shifts in dMRI metric distributions. [Top] AD subjects have a 1.02 SD increase in FW within the anterior commissure. [Bottom] TBI patients show a 1.31 SD decrease in AFD within the left IFOF.}
\label{fig:pathology_shift_examples}
\end{figure}

For instance, as illustrated in Figure~\ref{fig:pathology_shift_examples}, patients with AD typically exhibit elevated FW in certain white matter regions~\cite{Ji2017}, while individuals with TBI show altered AFD values~\cite{Wallace2020HBM}. This figure shows the standardized mean difference between patients and controls.
\begin{equation}
\label{eq:SD}
SD = \frac{\mu_{\text{patients}} - \mu_{\text{controls}}}{\sigma_{\text{controls}}}.
\end{equation}

This index quantifies how far, on average, the patient group lies from the control group in standardized units. These findings confirm that pathological subjects may occupy distinct regions of the diffusion metric space compared to healthy controls, reflecting disease-related variability that should be accounted for during harmonization.

\subsection*{Impact of Outliers on Harmonization}

When pathological subjects are included in ComBat parameter estimation, their abnormal diffusion values bias the site means and variances. Since ComBat assumes most samples follow comparable biological distributions, disease effects may be misinterpreted as scanner-related variability, leading to biased correction and suboptimal alignment of healthy subjects.

As illustrated in Figure~\ref{fig:clinicalcontext}, a site containing shifted pathological cases yields inflated mean and variance estimates, causing overcorrection of healthy data and resulting in residual misalignment of healthy controls across sites after harmonization. After correction, only the most extreme TBI cases remain distinguishable, while moderate cases overlap with healthy controls, compromising diagnosis using diffusion metrics.

\section*{Methods}
\subsection*{ComBat Overview}

ComBat \cite{Fortin2017} models site-related variability using an empirical Bayes framework that adjusts both the mean and variance of the data across sites.  For a step-by-step mathematical description of ComBat, please refer to~\cite{Jodoin2025}.
Let $y_{ijv}$ denote the value of a diffusion-derived feature (e.g., FA) for a brain region $v$, subject $j$, and site $i$. The data formation model assumes:
\begin{equation}
\label{eq:combat}
y_{ijv} = \alpha_v + \mathbf{x}_{ij}^\top \boldsymbol{\beta}_v + \gamma_{iv} + \delta_{iv} \, \varepsilon_{ijv}, 
\quad \varepsilon_{ijv} \sim \mathcal{N}(0, \sigma_v^2)
\end{equation}
where $\alpha_v$ is the global mean of the entire population across every site, $\mathbf{x}_{ij}$ represents biological covariates (e.g., sex and age) with coefficients $\boldsymbol{\beta}_v$, $\gamma_{iv}$ and $\delta_{iv}$ are the additive and multiplicative site effects, and $\varepsilon_{ijv}$ is Gaussian noise with a brain region-specific variance $\sigma_v^2$.  

The harmonization process follows three main steps.  
First, the effects of biological covariates (\( \hat{\alpha}_v, \hat{\boldsymbol{\beta}}_v \)) are regressed out, and the data are standardized within each site to remove differences in location and scale. This step isolates scanner-related variability by centering and scaling each feature according to the estimated site-specific mean and variance:
\begin{equation}
\label{eq:combat_standard}
z_{ijv} = \frac{y_{ijv} - \hat{\alpha}_v - \mathbf{x}_{ij}^\top \hat{\boldsymbol{\beta}}_v}{\hat{\sigma}_v}
\end{equation}
where $\hat{\sigma}_v$ corresponds to the residual standard deviation estimated from the ordinary least squares fit after removing the global mean, biological covariates, and site effects.

Second, the standardized parameters (\( \gamma_{iv}^\ast, \delta_{iv}^\ast \)) are estimated using an empirical Bayes framework, which assumes shared prior distributions across voxels,
\begin{equation}
\label{eq:zijv}
\gamma_{iv} \sim \mathcal{N}(\gamma_{i}, \tau_{i}^2), \qquad 
\delta_{iv}^{2} \sim \mathrm{InverseGamma}(\lambda_i, \theta_i),
\end{equation} 
allowing information to be pooled across features and stabilizing parameter estimation, particularly for sites with limited sample sizes.  
Finally, the additive and multiplicative site effects are removed, the standardized data are brought back to their original scale, and the covariate effects are reintroduced to obtain harmonized values:
\begin{equation}
\label{eq:ycombat}
y_{ijv}^{\text{ComBat}}  = \frac{\hat{\sigma}_v}{\hat{\delta}_{iv}^*} \left( z_{ijv} - \hat{\gamma}_{iv}^* \right) + \hat{\alpha}_v + \mathbf{x}_{ij}^\top \boldsymbol{\beta}_v.
\end{equation}

Although this framework effectively mitigates scanner-related variability, it assumes Gaussian residuals and homogeneous data within each site. These assumptions make it sensitive to pathological yet non-Gaussian outliers illustrated in the previous section.

\subsection*{Outlier Detection Methods}
\label{subsec:baselines}

In this paper, we tested ten outlier detection strategies to evaluate how effective they are at removing atypical subjects before harmonization, thereby ensuring unbiased estimation of site-related effects. 
All methods were applied after regressing out covariate effects (age, sex, handedness) to isolate site-related variability (see Eq.~(\ref{eq:combat_standard})). 

Each method operates either at the bundle level (univariate detection) or at the subject level (aggregating across all bundles and metrics). For each method, we empirically tested several thresholds, and the one that maximized harmonization performance on a validation set was retained.

\subsubsection*{Bundle-level Outlier Detection}

The following methods identify outliers independently for each \textit{bundle-metric} distribution. 

\paragraph{Z-score (ZS)~\cite{Iglewicz1993Outliers}}
ZS is a common outlier rejection method.  It uses a standard z-score as a parametric baseline:
\begin{equation}
\label{eq:zs}
ZS_{ijv} = \left| \frac{z_{ijv} - \mu_{iv}}{\sigma_{iv}} \right|,
\end{equation}
where \(\mu_{iv}\) and \(\sigma_{iv}\) denote the mean and standard deviation of \(z_{iv}\) across subjects \(j\) for metric-bundle \(v\) within site \(i\). Any points exceeding a certain threshold $T$ (in our case $T=3.0$) are considered statistically unlikely under a Gaussian assumption and are therefore discarded.

\paragraph{Interquartile Range (IQR)~\cite{Tukey1977}}
Outliers are defined as data points lying outside the interquartile interval:
\begin{equation}
z_{ijv} < Q_1 - k \cdot \text{IQR} \quad \text{or} \quad z_{ijv} > Q_3 + k \cdot \text{IQR},
\end{equation}
where $Q_1$ and $Q_3$ are the first and third quartiles of the distribution, and the IQR is defined as:
\begin{equation}
\text{IQR} = Q_3 - Q_1.
\end{equation}
A multiplier of $k = 1.5$ was empirically selected as the optimal threshold.

\paragraph{Median Absolute Deviation (MAD)~\cite{Leys2013}}
Outliers are detected using the modified z-score:
\begin{equation}
\label{eq:mad}
ZS_{ijv}^* = 0.6745 \cdot \frac{z_{ijv} - \text{median}_{k\in J_i}(z_{ikv})}{\text{MAD}_{iv}},
\end{equation}
where $J_i$ denotes the set of patients acquired at site $i$ and ${\text{MAD}_{iv}}$ is:
\begin{equation}
\text{MAD}_{iv} = \text{median}_{j\in J_i}\big( |z_{ijv} - \text{median}_{k\in J_i}(z_{ikv})| \big).
\end{equation}
All points with $ZS_{ijv}^* > T$ are excluded, where the threshold $T = 3.5$ was empirically determined.

\paragraph{Rousseeuw-Croux Estimators (Sn, Qn)~\cite{Rousseeuw1993}}
The robust dispersion estimators of Rousseeuw and Croux offer alternatives to the standard deviation:
\begin{align}
\text{Sn} &= 1.1926 \cdot \text{median}_{j\in J_i} \big( \text{median}_{k\in J_i} |z_{ijv} - z_{ikv}| \big), \\
\text{Qn} &= 2.2219 \cdot Q_1\text{ of} \Bigl\{|z_{ijv} - z_{ikv}| \Big| j,k \in J_i,\; j < k\Bigr\}.
\end{align}
Outliers are identified when:
\begin{equation}
\frac{|z_{ijv} - \text{median}_{k\in J_i}(z_{ikv})|}{\text{Sn}} > T
\end{equation}
or:
\begin{equation}
\frac{|z_{ijv}- \text{median}_{k\in J_i}(z_{ikv})|}{\text{Qn}} > T
\end{equation}
where $T = 3.0$ was set following cross-validation.

\paragraph{Mean-Median Shift (MMS)} 
Because pathological effects consistently shift diffusion metrics in one direction (e.g., increased FW, decreased AFD; Figs.~\ref{fig:clinicalcontext} and \ref{fig:pathology_shift_examples}), they distort the otherwise approximately Gaussian distribution. This property can be exploited through an iterative procedure that removes samples from the pathological tail until convergence.

The pathological direction determines which tail is trimmed, based on prior knowledge of how each metric behaves in disease. For instance, TBI subjects exhibit lower AFD values than the normative population, leading to removal of left-tail outliers relative to the population median. More generally, metrics known to increase with pathology, such as MD, RD, and FW, require right-tail trimming, whereas metrics that decrease with pathology, such as FA and AFD, require left-tail exclusion.

This MMS procedure is similar to the iterative truncated mean (ITM) filter~\cite{jiang2012itm}, but differs in two fundamental ways. First, instead of \emph{truncating} extreme values toward a threshold, MMS \emph{removes} data points entirely. Second, MMS only acts on the pathological side of the distribution. The iteration stops once the mean and median converge according to
\begin{equation}
\frac{|\mu_{iv} - \text{median}_{j\in J_i}(z_{ijv})|}{\text{median}_{j\in J_i}(z_{ijv})} < 10^{-3}.
\label{eq:mms_stop}
\end{equation}

\paragraph{Variance Symmetry (VS)} 
The VS heuristic follows the same trimming strategy as MMS but uses a different stopping criterion. It iteratively removes pathological-tail outliers until the distribution becomes symmetric around the median, stopping when the average absolute deviation on one side matches that of the other.
\begin{equation}
|\mu_{iv\_\text{left}} - \mu_{iv\_\text{right}}| \approx 0.
\label{eq:vs_symmetry}
\end{equation}

where
\begin{equation}
\mu_{iv\_\text{left}} = \frac{1}{N_{\text{left}}} \sum_{z_{ikv} < \text{median}_{j\in J_i}(z_{ijv})} |z_{ikv} - \text{median}_{j\in J_i}(z_{ijv})|,
\end{equation}

\begin{equation}
\mu_{iv\_\text{right}} = \frac{1}{N_{\text{right}}} \sum_{z_{ikv} > \text{median}_{j\in J_i}(z_{ijv})} |z_{ikv} - \text{median}_{j\in J_i}(z_{ijv})|.
\end{equation}

\subsubsection*{Subject-level Detection Methods}

These methods aggregate information across all bundles and metrics to identify subjects showing global or distributed abnormalities.

\paragraph{Global Z-score (G\_ZS)} 
For each subject $j$, we compute the mean absolute z-score across all bundles and metrics:
\begin{equation}
\overline{ZS}_{ij} = \frac{1}{V} \sum_{v=1}^{V} |ZS_{ijv}|
\end{equation}
where $ZS_{ijv}$ is defined at Eq.~(\ref{eq:zs}) and V is the number of different features (bundle–metric combinations).  Any subject with $\overline{ZS}_{ij} > 1.5$ is marked as an outlier.

\paragraph{Global MAD (G\_MAD)} 
This method follows the same principle as the Global Z-score, but replaces the standard z-score with the modified z-score defined in Eq.~(\ref{eq:mad}). 
Outliers are identified when the subject's average absolute modified z-score exceeds $T = 3.5$.

\subsubsection*{Subject-Level MLP Outlier Detector}
\label{sec:mlp_outlier}

We also implemented a simple MLP classifier that takes as input all diffusion-derived measurements from a single subject and predicts whether the subject is an outlier or not. Each subject is represented by 430 features, corresponding to 43 bundles and 10 diffusion metrics per bundle. The model outputs a single probability reflecting the likelihood of being abnormal relative to the site distribution.

The network consists of three fully connected hidden layers with 256, 128, and 64 neurons, respectively. Each layer is followed by batch normalization, ReLU activation, and dropout with a rate of 0.5. The output layer is a single linear neuron with sigmoid activation to produce a probability of being an outlier. 

\begin{figure*}
    \centering
    \includegraphics[width=0.99\linewidth]{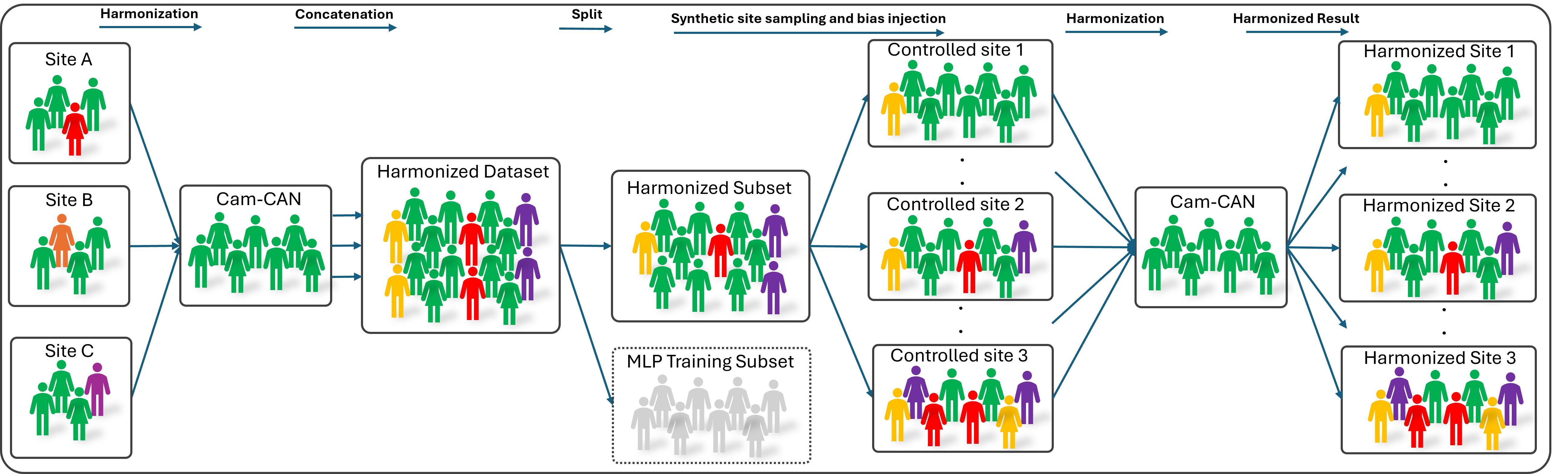}
    \caption[Data assembly pipeline]{
    (1) All available sites are first harmonized toward the CamCAN reference to construct a unified harmonized dataset.
    (2) The harmonized data are then split into two independent subsets, one used to train the MLP-based outlier detector and the other used to evaluate harmonization performance.
    (3) Control sites are generated by sampling subjects into multiple sites with varying proportions of diseased patients.\vspace{-0.2cm}
    }
    \label{fig:data_assembly}
\end{figure*}

\section*{Experiments}
\subsection*{Data and Preprocessing}

Data used in the preparation of this article were obtained from the Alzheimer’s Disease Neuroimaging Initiative (ADNI) database (adni.loni.usc.edu). ADNI was launched in 2003 as a public-private partnership, led by Principal Investigator Michael W. Weiner, MD. The primary goal of ADNI has been to test whether serial magnetic resonance imaging (MRI), positron emission tomography (PET), other biological markers, and clinical and neuropsychological assessments can be combined to measure the progression of mild cognitive impairment (MCI) and early Alzheimer’s disease (AD).

In addition to the ADNI dataset~\cite{Jack2008ADNIMRI,Petersen2010ADNIClinical}, diffusion MRI data were gathered from four other independent cohorts. The Transforming Research and Clinical Knowledge in Traumatic Brain Injury (TRACK--TBI) project contributed HC and subjects with mild to moderate TBI~\cite{TRACKTBI}. The UCLA Consortium for Neuropsychiatric Phenomics LA5c Study (LA5c) provided HC and participants with bipolar disorder (BIP), attention-deficit/hyperactivity disorder (ADHD), and schizophrenia (SCHZ)~\cite{Poldrack2016LA5c}. The SchizConnect (Mind Research Network) database provided HC and subjects with SCHZ~\cite{Wang2016SchizConnect}.  The last dataset is the Cambridge Centre for Ageing and Neuroscience (CamCAN) which provides healthy controls (HC) used in this paper as a normative reference~\cite{Shafto2014CamCAN,Taylor2017CamCAN}. 

The diffusion MRI datasets were processed using the TractoFlow pipeline~\cite{Theaud2020TractoFlow}. Diffusion-weighted volumes with b-values below $1200\,\mathrm{s/mm^2}$ were used to compute the standard DTI scalar maps, while volumes with b-values above $700\,\mathrm{s/mm^2}$ were employed to estimate the fiber orientation distribution function (fODF) metrics. The fODF was reconstructed using a spherical harmonics order of 8 and a common fiber response function across all subjects $(15, 4, 4) \times 10^{4}\,\mathrm{s/mm^2}$~\cite{Descoteaux2008Tractography}. All resulting metric maps were subsequently registered to the Montreal Neurological Institute (MNI) standard space. From these registered maps, the IIT Human Brain Atlas (v5.0;~\cite{Qi2021Regionconnect}) was used to extract the mean values of 10 diffusion metrics within 43 major white matter fiber bundles.

\subsection*{Construction of the Harmonized Dataset}

The selected datasets cover a broad range of neurological and psychiatric conditions, aiming to recreate a realistic clinical environment where sites have acquisition biases and patients present heterogeneous pathologies in varying proportions.

As illustrated in Figure~\ref{fig:data_assembly}, we created such an environment by first building a large harmonized dataset including multiple pathologies to simulate a realistic clinical setting (the leftmost A,B,C sites stand for the raw data from the aforementioned datasets). The raw data were harmonized with Pairwise-ComBat~\cite{Jodoin2025} toward CamCAN which acts as a normative reference. Among the 66 sites available, only those including at least 8 HC and 5 patients with a pathology were retained to ensure reliable harmonization. After filtering, 33 sites were kept, resulting in a dataset including 885 HC, 378 MCI, 200 AD, 208 TBI, 111 SCHZ, 48 BIP, and 39 ADHD for a total of 1,869 participants (aka the "harmonized dataset" in Figure~\ref{fig:data_assembly}).

The harmonized dataset was then divided into two equal parts: one part to evaluate the harmonization quality and the other to train the MLP.  Each subset was augmented (approximately tripled) by applying small perturbations to the data to increase variability, reduce overfitting, and prevent contamination between training and validation samples.

\subsection*{Control Site Generation}

From the harmonized subset, we generated a set of control sites with various disease prevalence and acquisition biases. 100 subjects were sampled from the harmonized subset with the following proportion of diseases: 3\%, 10\%, 30\%, 50\%, 70\%, or 80\%. For each prevalence level, 40 independent control sites were created, resulting in a total of 240 sites. Diseased subjects could come from any of the pathologies present in the pooled dataset (AD, MCI, TBI, SCHZ, BIP, ADHD). To simulate realistic site effects, we injected a site-wise artificial bias into each sampled control cohort. 
For a subject $j$ from site $i$ and feature $v$, the artificial bias was applied by first removing the effect of biological covariates, then applying multiplicative and additive site effects, and finally reintroducing the covariate contribution:
\begin{equation}
\tilde{y}_{ijv}
=
\alpha_v
+
\mathbf{x}_j^\top \boldsymbol{\beta}_v
+
\textcolor{red}{\gamma_{iv}}
+
\textcolor{red}{\delta_{iv}}
\big(y_{ijv} - \alpha_v - \mathbf{x}_j^\top \boldsymbol{\beta}_v\big),
\label{eq:synthetic_bias}
\end{equation}
where $y_{ijv}$ denotes the original feature value, $\alpha_v$ is the global mean across sites, $\mathbf{x}_j$ represents biological covariates (e.g., age, sex) with associated coefficients $\boldsymbol{\beta}_v$, and $\textcolor{red}{\gamma_{iv}}$ and $\textcolor{red}{\delta_{iv}}$ correspond to the synthetic additive and multiplicative site effects, respectively. This artificial bias injection can be interpreted as the inverse operation of the ComBat harmonization step described in Eq.~(\ref{eq:ycombat}).

With the control sites at hand, the experiments consisted to harmonizing these sites onto CamCAN and measuring how well the various filtering methods compensate for the presence of diseased subjects. 

\subsection*{MLP Training}
The MLP was trained to detect subject-level outliers by learning deviations relative to site-specific distributions rather than absolute diffusion values. The dataset was split 80/10/10 for training, validation, and testing, while preserving the proportion of healthy and pathological subjects across all splits.

To enhance robustness and expose the model to varied clinical acquisition settings, synthetic sites were generated from the training subset using the same procedure as control sites, grouping 100 subjects with varying disease ratios. Features were standardized within each cohort using z-scores per bundle and metric, ensuring the model learned local deviations relative to its site distribution rather than relying on global intensity differences that may be confounded by site effects. Subjects were then presented independently to the network during training.

The model was optimized using the Adam optimizer. Hyperparameters were selected using Optuna-based optimization, and training was performed with mini-batches of 64 subjects using a binary cross-entropy loss with logits. To avoid degrading the harmonization process by incorrectly excluding healthy subjects, a higher penalty was assigned to misclassified healthy controls, encouraging conservative predictions.

To gauge the performances of the MLP outlier rejection method on real multi-site data, we adopted a bootstrap-based site-level resampling strategy. At each bootstrap iteration, three sites were randomly selected and held out, one from ADNI~\cite{Jack2008ADNIMRI,Petersen2010ADNIClinical}, one from LA5c~\cite{Poldrack2016LA5c}, and one from Track-TBI~\cite{TRACKTBI}. These held-out sites were excluded from the MLP training and treated as independent test sites. The MLP trained on the remaining sites was then applied to the held-out sites and directly compared to the NO\_FILTERING baseline. This entire procedure was repeated over 30 bootstrap iterations.

\subsection*{Evaluation Metric}
To assess harmonization quality, we use the standardized Mean Absolute Error (STD\_MAE). 
The MAE is computed as the average absolute difference between each subject’s harmonized feature value 
and its reference value from the harmonized subset. It is then normalized by the standard deviation 
of the corresponding feature in the harmonized dataset, yielding a scale-independent measure that 
enables fair comparisons across bundles and metrics. This evaluation is only possible when using a fixed reference site, which is why we rely on ComBat adaptations such as Pairwise-ComBat to determine whether biased data return to their expected location after re-harmonization.

We also used the Bhattacharyya distance, a method proposed by Girard et al.~\cite{Girard2025} to measure how much the distribution of two healthy populations align on each other after harmonization. This metric is applied on real sites (not control sites).

\section*{Results}
\begin{figure*}[!t]
    \centering
    \includegraphics[width=0.99\linewidth]{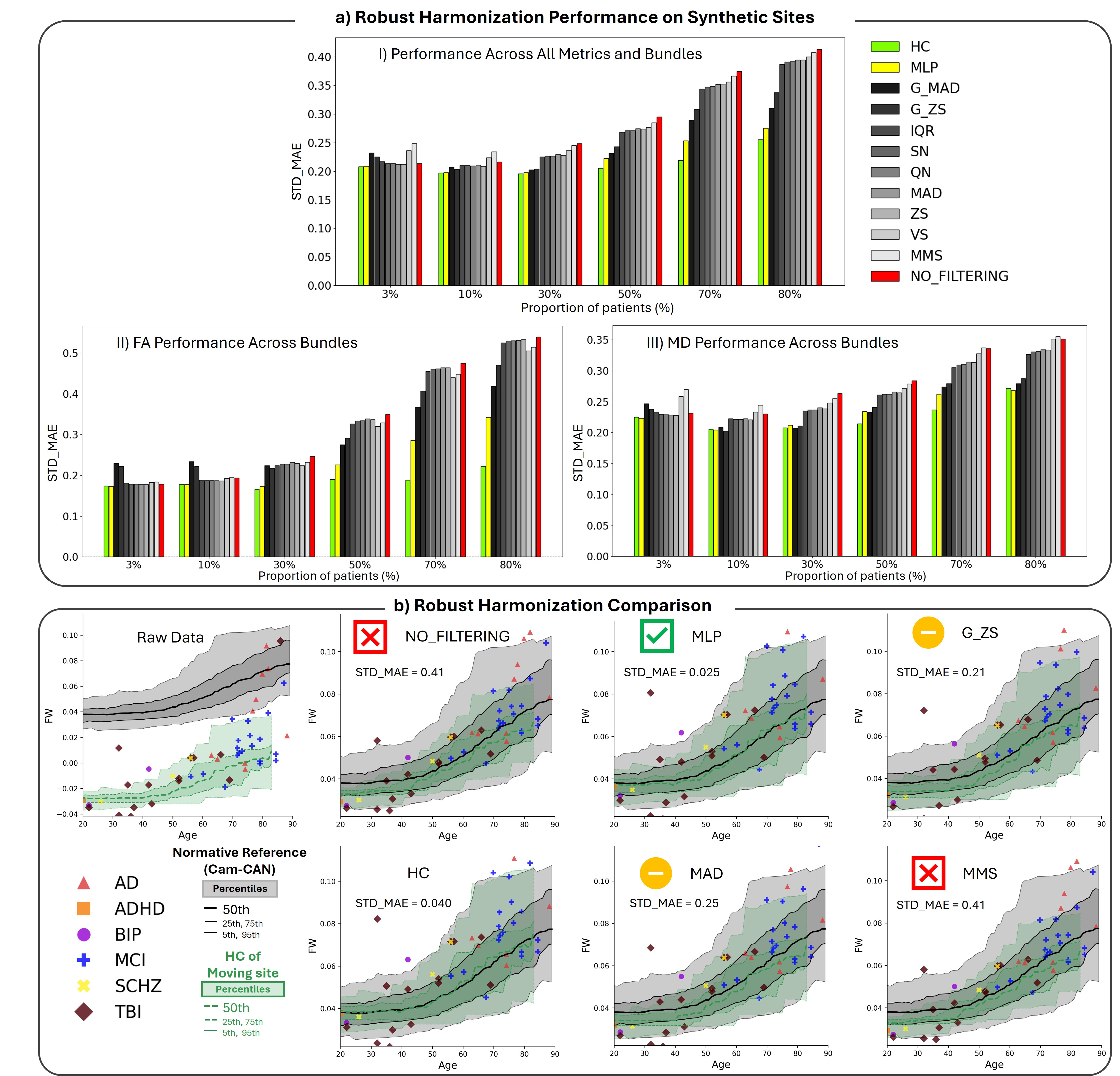}
    \caption{
        (a) Global harmonization performance. 
        (I) Mean STD-MAE averaged across all diffusion metrics and bundles. 
        (II) STD-MAE computed for FA across bundles. 
        (III) STD-MAE computed for MD across bundles. 
        As the proportion of pathological subjects increases, classical statistical approaches exhibit increasing error, whereas MLP-based filtering remains more stable and closer to the HC baseline.  
        (b) Illustration of outlier-handling strategies prior to harmonization for the FW metric in the right uncinate fasciculus (UF right) within a site containing 50\% pathological subjects, shown as a function of age. Raw data, no filtering, HC-only filtering, MLP-based filtering, and statistical approaches (G\_ZS, MAD, MMS) are displayed.
        }
    \label{fig:mae_examples_pairwise}
\end{figure*}

\subsection*{Control Sites: Outlier Rejection Methods}

Figure~\ref{fig:mae_examples_pairwise}(a) summarizes the performance of the different outlier rejection strategies for sites with 100 patients harmonized with Pairwise-ComBat. The green bar corresponds to the harmonization error obtained when using only HC, which represents the ideal reference scenario. The red curve shows the error obtained without any outlier rejection, the yellow bar corresponds to the MLP, and the remaining nine bars represent the various filtering methods.

Panel (I) reports the average performance over all diffusion metrics and white-matter bundles, while panels (II) and (III) focus specifically on the FA and MD metrics, respectively.

At first glance, when the proportion of diseased subjects is low (below 30\%), most filtering strategies fail to meaningfully outperform the NO\_FILTERING baseline. However, for prevalence levels that are more representative of routine clinical settings (30\% and above), all outlier rejection methods outperform NO\_FILTERING. Overall, the MMS and VS methods exhibit weaker performance. This behavior can be attributed to the fact that the underlying distributions may remain close to Gaussian even in the presence of outliers, preventing these methods from being triggered effectively. Conversely, when only healthy controls are present, the distributions are not perfectly Gaussian, leading these methods to activate unnecessarily and to remove an excessive number of subjects. This effect explains the pronounced performance fluctuations observed at low prevalence levels (e.g., 3\% and 10\%).

\begin{table}[tp]
\centering
\begin{tabular}{c c | c c}
\toprule
\textbf{Disease ratio} &
\textbf{HC} &
\textbf{MLP} &
\textbf{NO\_FILTERING} \\
\midrule
3\%  & 0.29 & 0.31 & \textbf{0.30} \\
10\% & 0.28 & \textbf{0.29} & 0.30 \\
30\% & 0.28 & \textbf{0.29} & 0.33 \\
50\% & 0.31 & \textbf{0.32} & 0.40 \\
70\% & 0.34 & \textbf{0.35} & 0.56 \\
80\% & 0.41 & \textbf{0.39} & 0.64 \\
\bottomrule
\end{tabular}
\caption{Mean of the top 10\% highest STD\_MAE across disease ratios over all metric bundles (HC, MLP, and NO\_FILTERING)}
\label{tab:synthetic_sites_max}
\end{table}

Among the filtering-based approaches, G\_MAD and G\_ZS provide the most consistent improvements, particularly when the proportion of outliers is high. This highlights the benefit of leveraging information jointly across multiple metrics and bundles. Nevertheless, their average performance remains inferior to that of the MLP. Moreover, when the prevalence of diseased subjects is below 30\%, both G\_MAD and G\_ZS tend to noticeably degrade performance.

Overall, the MLP consistently outperforms all competing methods, offering substantial gains with limited adverse effects. Its advantage is especially pronounced for the FA metric, where classical and graph-based methods either fail to improve performance or even deteriorate it. In contrast, the MLP achieves substantially lower errors across all prevalence levels.

In Table~\ref{tab:synthetic_sites_max}, we look at the top 10\% highest STD\_MAE values, thereby characterizing worst-case harmonization outcomes. In low disease-ratio scenarios (3--10\%), the MLP filtering may occasionally induce slight degradations compared to NO\_FILTERING, reflecting situations where pathological outliers are scarce.
However, for disease ratios of 30\% and above, corresponding to the clinically relevant regimes targeted in this work, NO\_FILTERING exhibits a sharp increase in extreme errors, whereas the worst-case performance of the MLP remains close to that of HC. Interestingly, although Fig.~\ref{fig:mae_examples_pairwise}(a) shows that HC outperforms the MLP on average, their worst-case behaviors remain comparable.

Fig.~\ref{fig:mae_examples_pairwise}(b) illustrates a representative harmonization example on a control site containing 50\% diseased subjects, using the FW metric and the right uncinate fasciculus (UF\_R) bundle. The MLP yields an almost ideal correction, with the HC distribution (green) closely aligned with the normative reference. The MAD method produces an intermediate outcome: while overall alignment improves, several diseased subjects remain improperly merged with the healthy population. G\_ZS performs slightly better than MAD but still does not reach the performance of the MLP. In contrast, MMS performs poorly in this scenario, yielding results comparable to those obtained without any outlier rejection.

\subsection*{Sample Size Effect}
We repeated the evaluation with Pairwise-ComBat by varying the total number of subjects (20 to 60) and considering disease ratios of 50\%, 70\%, and 80\%, to determine the minimum number of subjects required for reliable MLP performance.

\begin{figure}
    \centering
    \includegraphics[width=0.99\linewidth]{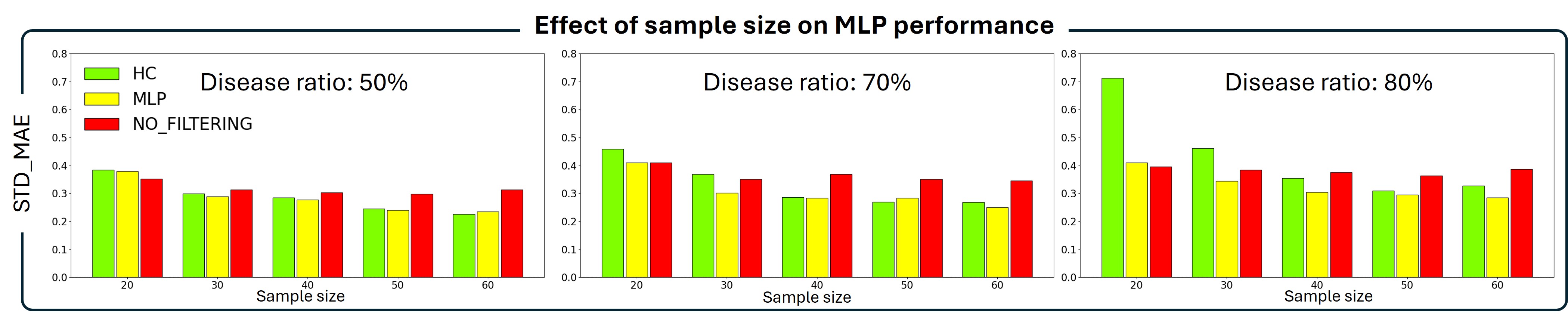}
    \caption{STD\_MAE measured at different disease ratios (50\%, 70\%, and 80\%) for MLP, HC, and NO\_FILTERING across increasing total numbers of patients (20-60).}
    \label{fig:mae_sizes}
\end{figure}

Figure~\ref{fig:mae_sizes} shows that HC-only harmonization becomes unreliable at small sample sizes and high disease ratios, with a sharp increase in error at 20 subjects for 70\% and 80\% due to an insufficiently large healthy subset (a 20 subjects site including 80\% outliers contains only 4 HC). Overall, HC provides consistent improvements only when at least 50 subjects are available.

At 20 subjects, the MLP yields limited gains. However, from 30 subjects onward, it consistently outperforms the NO\_FILTERING baseline, with the performance gap increasing as sample size grows.

\vspace{-0.2cm}
\subsection*{Comparison with ComBat Variants}
\begin{figure}[tp]
    \centering
    \includegraphics[width=0.99\linewidth]{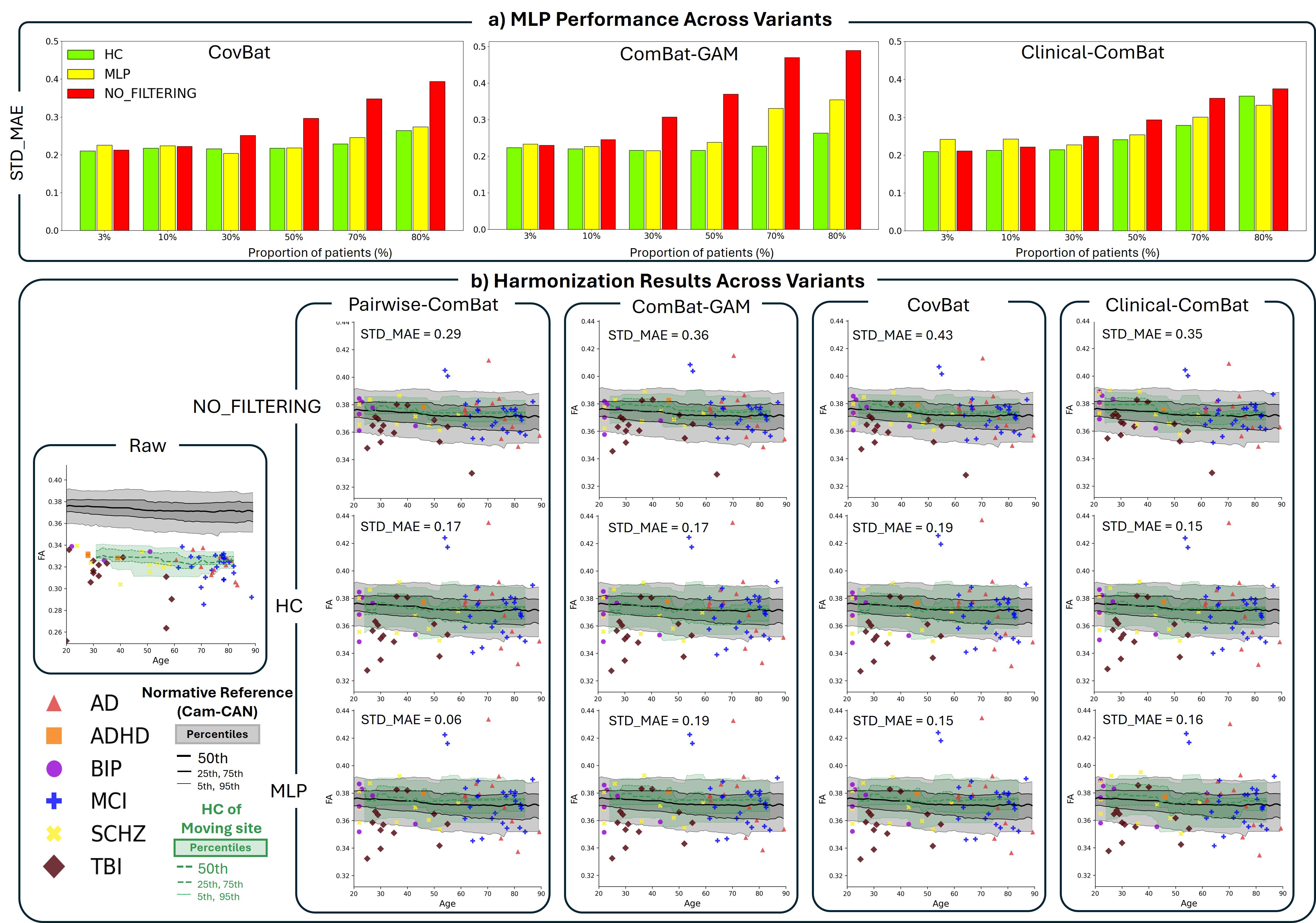}
    \caption{
    (a) STD-MAE across all metric bundles for CovBat, ComBat-GAM, and Clinical-ComBat under MLP, HC-only, and NO\_FILTERING.
    (b) Harmonization example for FA in the left STT bundle for a site containing 70\% pathological subjects under CovBat, ComBat-GAM, and Clinical-ComBat, using MLP, HC-only, and NO\_FILTERING.
    }
    \label{fig:mae_example_variants}
\end{figure}

We tested the best outlier rejection method (MLP) with three variants of ComBat, namely CovBat~\cite{Chen2021CovBat}, ComBat-GAM~\cite{Pomponio2020}, and Clinical-ComBat~\cite{Girard2025}. For consistency, CovBat and ComBat-GAM were modified similarly to Pairwise-ComBat by harmonizing each moving site to a normative reference population.  Results are presented in  Fig.~\ref{fig:mae_example_variants}(a).  All three methods are sensitive to diseased subjects. The consistent gap between NO\_FILTERING and HC harmonization confirms that pathological contamination degrades site-effect estimation. This effect is especially pronounced for CovBat and ComBat-GAM, where performance drops substantially. In these cases, the MLP markedly reduces the error, bringing it closer to the HC setting. Clinical-ComBat yields less pronounced average improvements, yet still offers meaningful gains at higher disease proportions. Moreover, it remains more stable when all available data are used for harmonization. For example, at 80\% pathology, its degradation reaches approximately 0.35, whereas ComBat-GAM rises to nearly 0.5 when outliers are left in the data.

Fig.~\ref{fig:mae_example_variants}(b) reports the harmonization results for a control site containing 70\% diseased subjects. The impact of contamination becomes obvious for all ComBat variants as the NO\_FILTERING results have a much larger STD\_MAE. Without filtering, many pathological subjects overlap with the normative population, whereas this overlap disappears when only HC are used. Across all four ComBat variants, adding the MLP produces almost perfectly aligned HC populations (the green plot aligned with the gray plot), where diseased patients remain clearly separated from the normative distribution.

\subsection*{Performance on Real Sites} 
As shown in Table~\ref{tab:real_sites}, the proposed MLP improves the Bhattacharyya distance relative to the NO\_FILTERING baseline for nearly all metrics. For AD, ADT, FW, MD, MDT, and RD, improvements reach at least 24\%, up to 36\% for AD. FAT is the only metric showing a negligible deterioration.

\begin{table}[h!]
\centering
\begin{tabular}{c c c c}
\hline
Metric & Raw & No Filtering & \textbf{MLP} \\
\hline
ad  & 0.50 & 0.036 & \textbf{0.023} \\
adt & 0.58 & 0.032 & \textbf{0.022} \\
afd & 0.45 & 0.026 & \textbf{0.025} \\
fa  & 0.24 & 0.021 & \textbf{0.020} \\
fat & 0.38 & \textbf{0.020} & 0.021 \\
fw  & 0.33 & 0.036 & \textbf{0.027} \\
md  & 0.26 & 0.030 & \textbf{0.020} \\
mdt & 0.21 & 0.025 & \textbf{0.019} \\
rd  & 0.19 & 0.025 & \textbf{0.018} \\
rdt & 0.13 & 0.021 & \textbf{0.019} \\
\hline
\end{tabular}
\caption{Bhattacharyya Distance for Harmonized Data Across Real Sites (Raw, No Filtering, MLP Filtering) for the Axial diffusivity (ad), the apparent fiber density (afd), the fractional anisotropy (FA), mean diffusivity (md) and the radial diffusivity (rd).  adt, fat, mdt and rdt are freewater corrected metrics.}
\label{tab:real_sites}
\end{table}

\section*{Discussion and Recommendations}
Overall, the proposed filtering strategies yield measurable improvements across harmonization methods, but the magnitude and reliability of these gains depend primarily on the proportion of pathological subjects. Although the exact contamination level is unknown in practice, cohorts can generally be positioned within two broad regimes: low contamination, typically below 30\%, and clinically realistic high contamination, where the proportion of patients exceeds 30\%. At least 30 total patients are required for outlier rejection to be stable and beneficial. Below this threshold, filtering becomes unreliable and should be avoided.

When the proportion of patients is low, filtering is not critical and gains remain limited. This scenario, which is not the primary target of this work, is more typical of control-enriched research cohorts than clinical practice. The MLP may provide slight improvements with minimal risk, but a conservative method such as IQR or MAD is generally sufficient to remove extreme deviations. In all cases, the rejection threshold can be increased to further reduce the risk of unintended deterioration.

The clinically relevant scenario corresponds to contamination levels of 30\% and above, as reasonably expected in routine diagnostic settings. In this regime, any outlier rejection strategy improves harmonization relative to NO\_FILTERING, but the MLP consistently provides the largest gains across harmonization variants. The MLP framework has the practical limitation of requiring the exact metrics and bundle configuration used during training. Consequently, if a dataset does not match this feature space, generalized robust approaches such as G\_MAD or G\_ZS provide suitable alternatives, as they provide substantial improvements without a tendency to degrade performance when contamination is elevated.

In summary, when harmonizing data using ComBat or any of its variants in an unlabeled population where patient status is unknown, below 30 subjects, aggressive filtering should be avoided. In low-contamination settings, conservative filtering such as IQR is sufficient. In clinically realistic high-contamination regimes with adequate sample size, the MLP approach should be preferred, with G\_MAD or G\_ZS serving as practical alternatives when the trained model cannot be applied. Across all scenarios, adjusting the rejection threshold offers a simple mechanism to balance robustness and conservativeness.
\vspace{-0.4cm}
\section*{Conclusion}
This work highlights a limitation of ComBat-based harmonization in clinical diffusion MRI: site-effect estimation is often biased by pathological cases that violate the model’s Gaussian assumptions. When healthy and diseased subjects are blindly mixed and harmonized toward a normative reference, ComBat may distort biological variability and misalign healthy controls. We introduce Robust-ComBat, which integrates ComBat with outlier filtering. In a large evaluation (10 filtering strategies, 4 ComBat variants, 6 neurological conditions), conventional statistical filters proved unstable in highly contaminated settings. In contrast, a lightweight MLP trained on multi-bundle, multi-metric diffusion profiles showed the most consistent performance, reducing harmonization error, limiting extreme failures at high disease prevalence, and preserving disease-related deviations. Robust-ComBat therefore supports clinically realistic workflows where local healthy cohorts are scarce and patient status is often unknown at acquisition.

Future work should expand training across more protocols, populations, and pathologies to enhance robustness and portability. Prospective validation in real clinical settings would better reflect routine diagnostic conditions. Finally, assessing the impact of Robust-ComBat on downstream tasks—such as normative modeling, anomaly detection, and disease classification—would clarify its practical clinical value beyond distributional alignment.

\vspace{-0.2cm}
\section*{Data and code availability}
\vspace{-0.2cm}
All materials necessary to reproduce the results presented in this study will be made openly accessible following acceptance of the manuscript. The dataset has been deposited on Zenodo and can be accessed at https://zenodo.org/records/18802256. The source code used for data processing, analysis, and figure generation is maintained in a public GitHub repository available at https://github.com/scil-vital/ComBat-Robust.

\bibliographystyle{unsrt}
\bibliography{arxiv_bib}

\section*{Acknowledgements}
Data collection and sharing for this project was provided by the Cambridge Centre for Ageing and Neuroscience (CamCAN). CamCAN funding was provided by the UK Biotechnology and Biological Sciences Research Council (grant number BB/H008217/1), together with support from the UK Medical Research Council and University of Cambridge, UK.

Data collection and sharing for this project was funded by the Alzheimer's Disease Neuroimaging Initiative (ADNI) (National Institutes of Health Grant U01 AG024904) and DOD ADNI (Department of Defense award number W81XWH-12-2-0012). ADNI is funded by the National Institute on Aging, the National Institute of Biomedical Imaging and Bioengineering, and through generous contributions from the following: AbbVie, Alzheimer’s Association; Alzheimer’s Drug Discovery Foundation; Araclon Biotech; BioClinica, Inc.; Biogen; Bristol-Myers Squibb Company; CereSpir, Inc.; Cogstate; Eisai Inc.; Elan Pharmaceuticals, Inc.; Eli Lilly and Company; EuroImmun; F. Hoffmann-La Roche Ltd and its affiliated company Genentech, Inc.; Fujirebio; GE Healthcare; IXICO Ltd.; Janssen Alzheimer Immunotherapy Research \& Development, LLC.; Johnson \& Johnson Pharmaceutical Research \& Development LLC.; Lumosity; Lundbeck; Merck \& Co., Inc.; Meso
Scale Diagnostics, LLC.; NeuroRx Research; Neurotrack Technologies; Novartis Pharmaceuticals
Corporation; Pfizer Inc.; Piramal Imaging; Servier; Takeda Pharmaceutical Company; and Transition
Therapeutics. The Canadian Institutes of Health Research is providing funds to support ADNI clinical sites
in Canada. Private sector contributions are facilitated by the Foundation for the National Institutes of Health
(www.fnih.org). The grantee organization is the Northern California Institute for Research and Education,
and the study is coordinated by the Alzheimer’s Therapeutic Research Institute at the University of Southern
California. ADNI data are disseminated by the Laboratory for Neuro Imaging at the University of Southern
California.

\section*{Consortia\\
Alzheimer’s Disease Neuroimaging Initiative}
Michael W. Weiner\textsuperscript{1},
Paul Aisen\textsuperscript{2},
Ronald Petersen\textsuperscript{3},
Clifford R. Jack Jr.\textsuperscript{3},
William Jagust\textsuperscript{4},
Susan Landau\textsuperscript{4},
Monica Rivera-Mindt\textsuperscript{5},
Ozioma Okonkwo\textsuperscript{6},
Leslie M. Shaw\textsuperscript{7},
Edward B. Lee\textsuperscript{7}
% ...

\newcommand\blfootnote[1]{%
  \begingroup
  \renewcommand\thefootnote{}\footnote{#1}%
  \addtocounter{footnote}{-1}%
  \endgroup
}
\blfootnote{
\textsuperscript{1} University of California, San Francisco; Northern California Institute for Research and Education, USA, 
\textsuperscript{2} University of Southern California, USA,
\textsuperscript{3} Mayo Clinic, Rochester, USA,
\textsuperscript{4} University of California, Berkeley, USA, 
\textsuperscript{5} Fordham University; Mt. Sinai Medical Center, USA, 
\textsuperscript{6} University of Wisconsin, USA, 
\textsuperscript{7} University of Pennsylvania, USA, 
\textsuperscript{8} University of California, Los Angeles, USA, 
\textsuperscript{9} University of California, Davis, USA, 
\textsuperscript{10} Harvard University, USA, 
\textsuperscript{11} Indiana University School of Medicine, USA, 
\textsuperscript{12} Washington University, St. Louis, USA, 
\textsuperscript{13} Eisai Pharmaceuticals, USA, 
\textsuperscript{14} University of Washington, USA, 
\textsuperscript{15} Northern California Institute for Research and Education, USA, 
\textsuperscript{16} University College London, United Kingdom, 
\textsuperscript{17} Mt. Sinai Medical Center, USA, 
\textsuperscript{18} University of Pittsburgh, USA, 
\textsuperscript{19} University of Michigan, USA, 
\textsuperscript{20} Duke University / AD Metabolomics Consortium, USA, 
\textsuperscript{21} University of California, Irvine, USA, 
\textsuperscript{22} Khachaturian, Radebaugh \& Associates (KRA), Inc, USA, 
\textsuperscript{23} General Electric, USA, 
\textsuperscript{24} University of Connecticut, USA, 
\textsuperscript{25} National Institute on Aging / NIH, USA, 
\textsuperscript{26} Alzheimer's Association, USA, 
\textsuperscript{27} National Institute of Mental Health, USA, 
\textsuperscript{28} Rush University, USA, 
\textsuperscript{29} University of California, San Diego, USA, 
\textsuperscript{30} Denali Therapeutics, USA, 
\textsuperscript{31} Massachusetts General Hospital, USA, 
\textsuperscript{32} Biogen, USA
}

\section*{Author contributions statement}

Y.D. conceived and designed the experiments, developed the proposed MLP-based method, structured and processed the data, wrote the code for the experiments, performed the analyses, produced all figures, and wrote the paper. P-M.J. supervised the project, wrote the paper and funded the work.

Data used in preparation of this article were partly obtained from the Alzheimer’s Disease Neuroimaging Initiative (ADNI) database (adni.loni.usc.edu). As such, the investigators within the ADNI contributed to the design and implementation of ADNI and/or provided data but did not participate in analysis or writing of this article.

\section*{Additional information}

\textbf{Competing interests} The authors declare no competing interests. 

\textbf{Funding}
This work was supported by the Natural Sciences and Engineering Research Council of Canada through its discovery grant program and the Arbour Foundation.
\input{supp}
\end{document}

%% file: supp.tex
\clearpage
\section*{Supplementary Material}

\setcounter{figure}{0}
\renewcommand{\thefigure}{S\arabic{figure}}

This document contains complementary results to those presented in the paper.

\paragraph{Figures S1 to S8 complement Figure 4 in the paper.} They report outlier detection performance for the remaining diffusion metrics (AD, ADT, AFD, FAT, FW, MDT, RD and RDT) not displayed in Figure 4.

\begin{figure}[ht]
    \centering
    \includegraphics[width=0.99\textwidth]{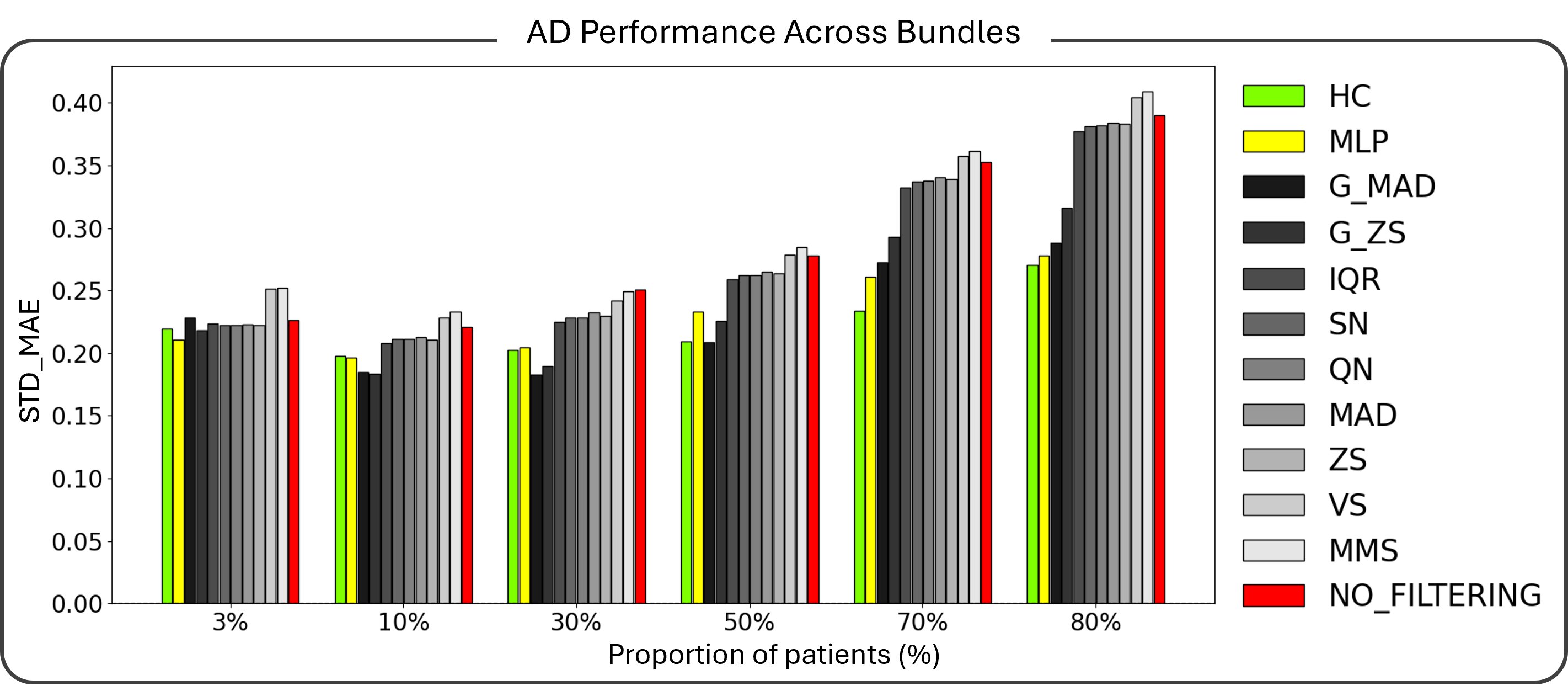} 
    \caption{This illustration is complementary to figure 3 in the paper.}
    \label{fig:S1}
\end{figure}

\begin{figure}[ht]
    \centering
    \includegraphics[width=0.99\textwidth]{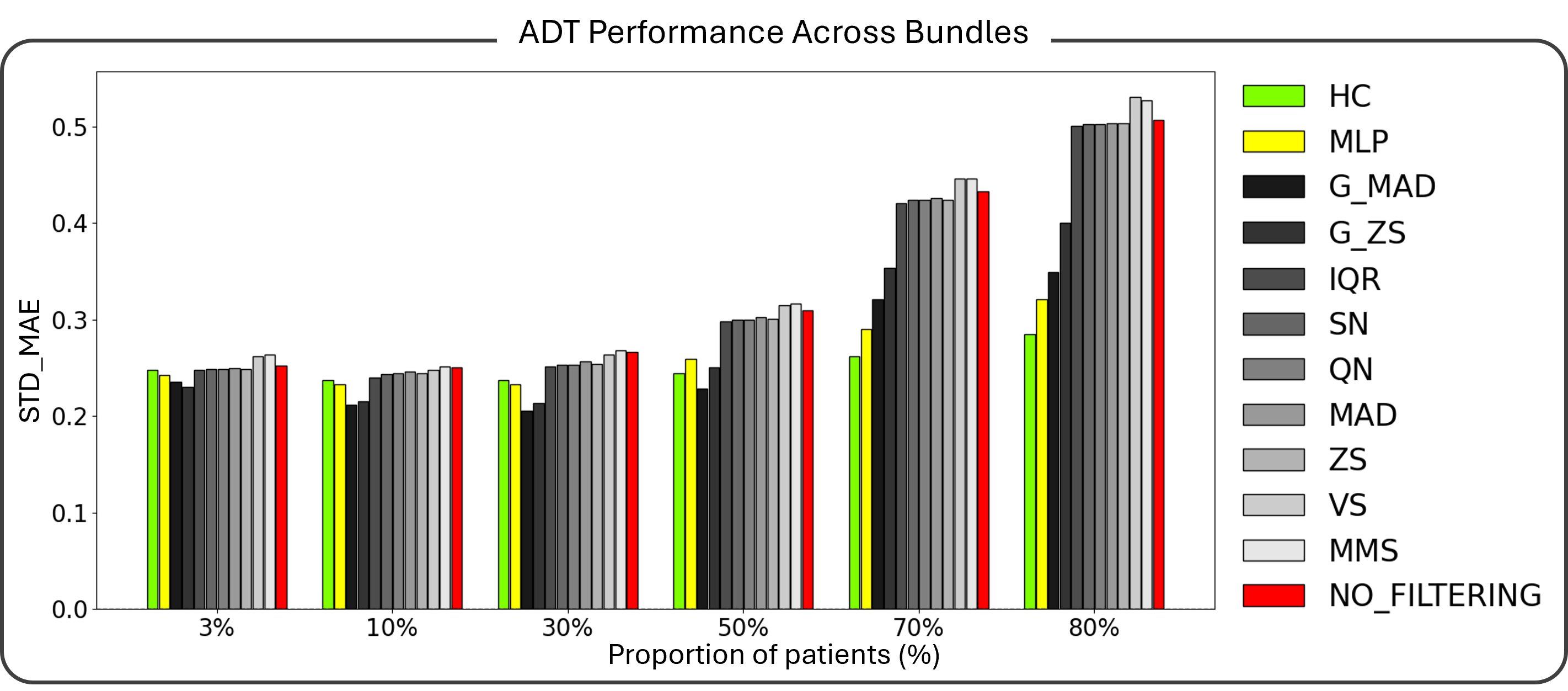} 
    \caption{This illustration is complementary to figure 3 in the paper.}
    \label{fig:S2}
\end{figure}

\begin{figure}[ht]
    \centering
    \includegraphics[width=0.99\textwidth]{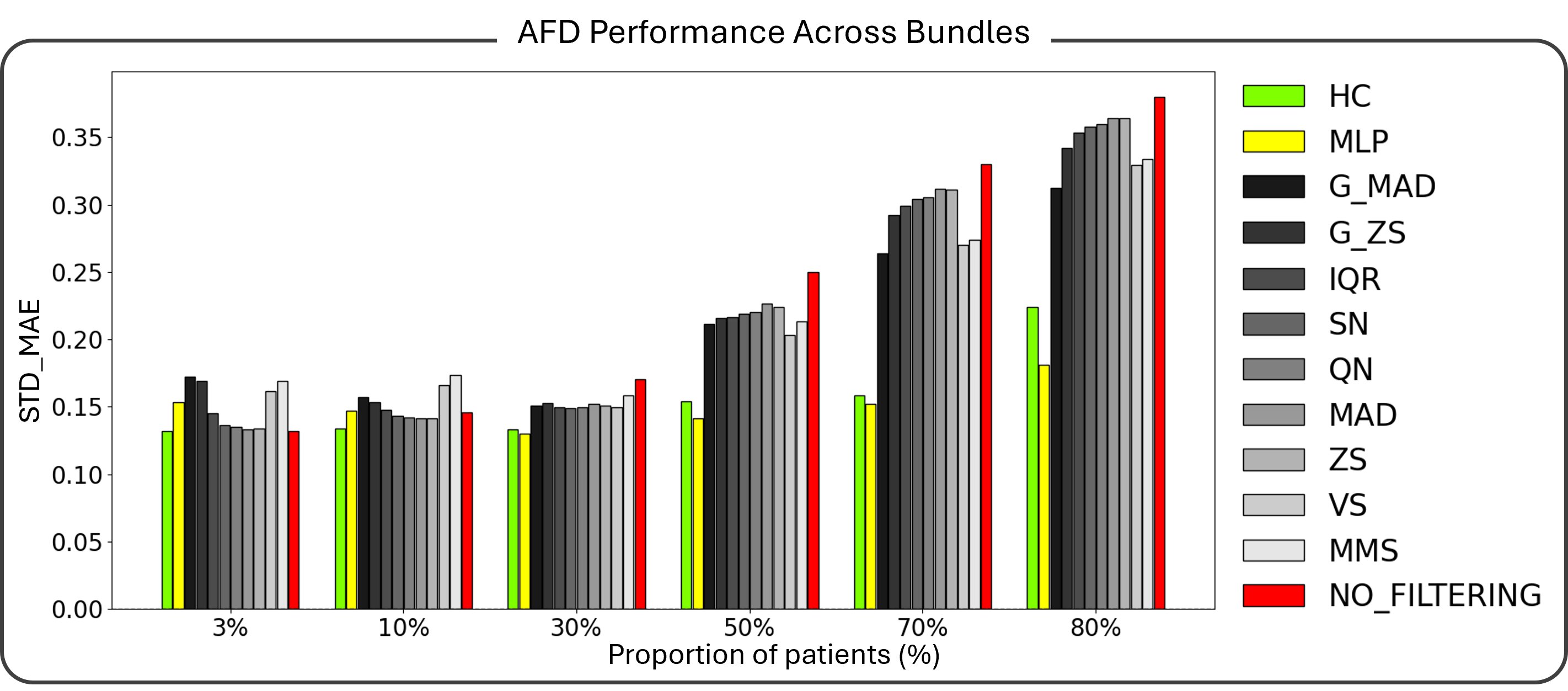} 
    \caption{This illustration is complementary to figure 3 in the paper.}
    \label{fig:S3}
\end{figure}

\begin{figure}[ht]
    \centering
    \includegraphics[width=0.99\textwidth]{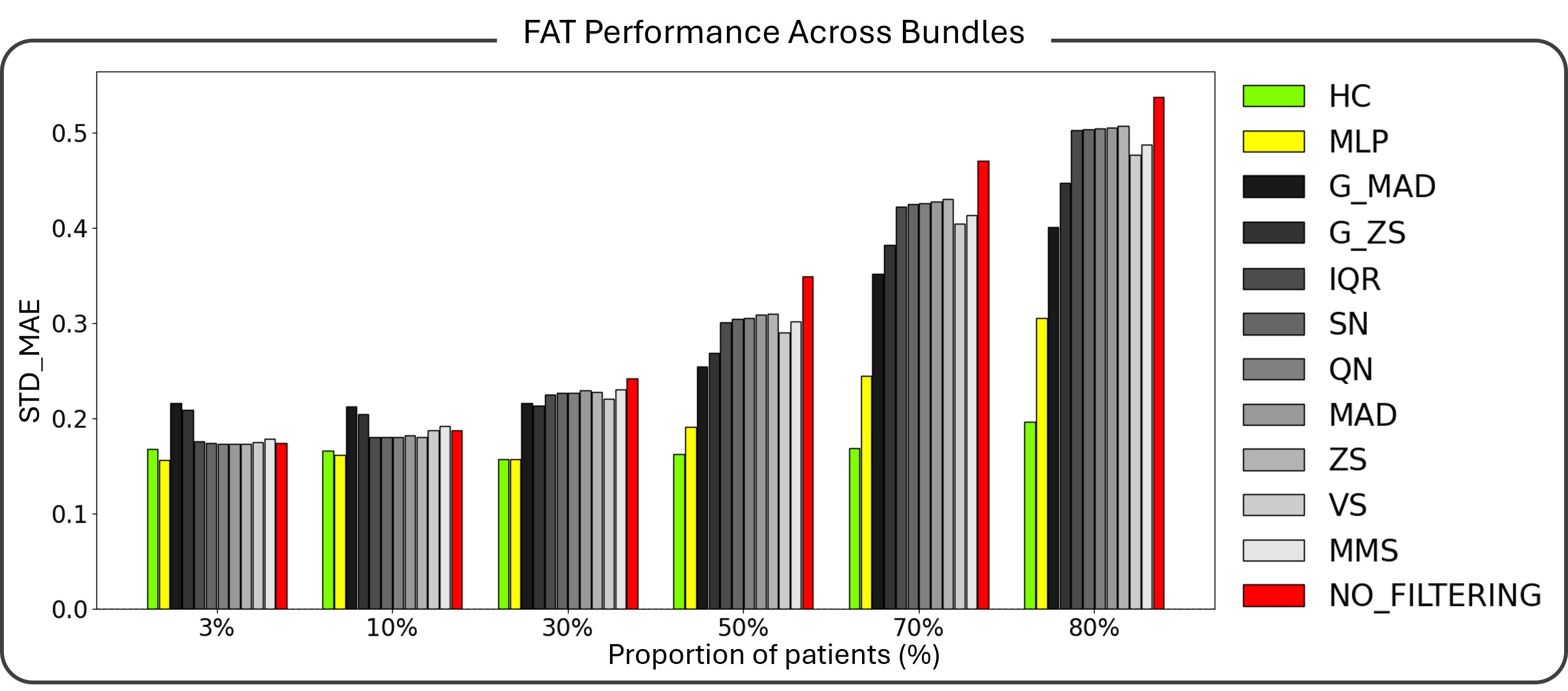} 
    \caption{This illustration is complementary to figure 3 in the paper.}
    \label{fig:S4}
\end{figure}

\begin{figure}[ht]
    \centering
    \includegraphics[width=0.99\textwidth]{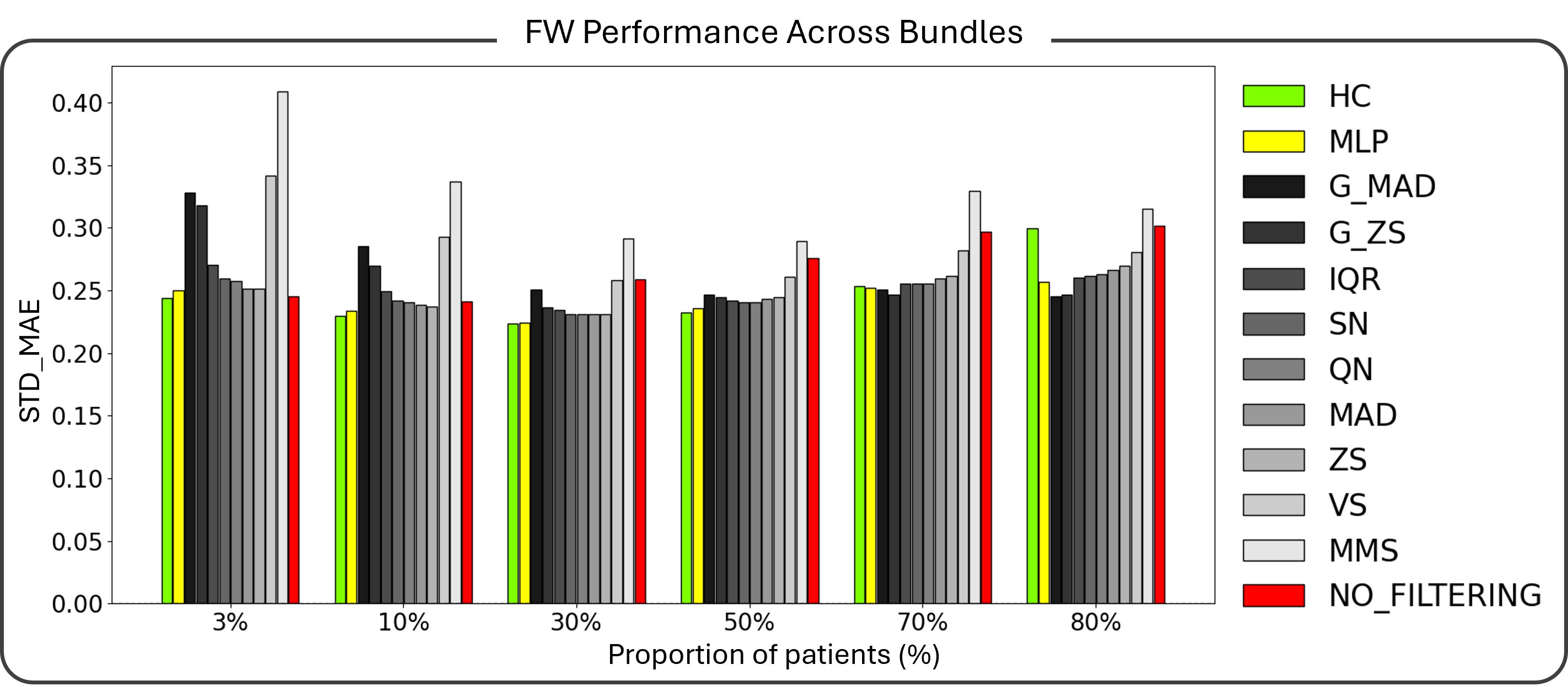} 
    \caption{This illustration is complementary to figure 3 in the paper.}
    \label{fig:S5}
\end{figure}

\begin{figure}[ht]
    \centering
    \includegraphics[width=0.99\textwidth]{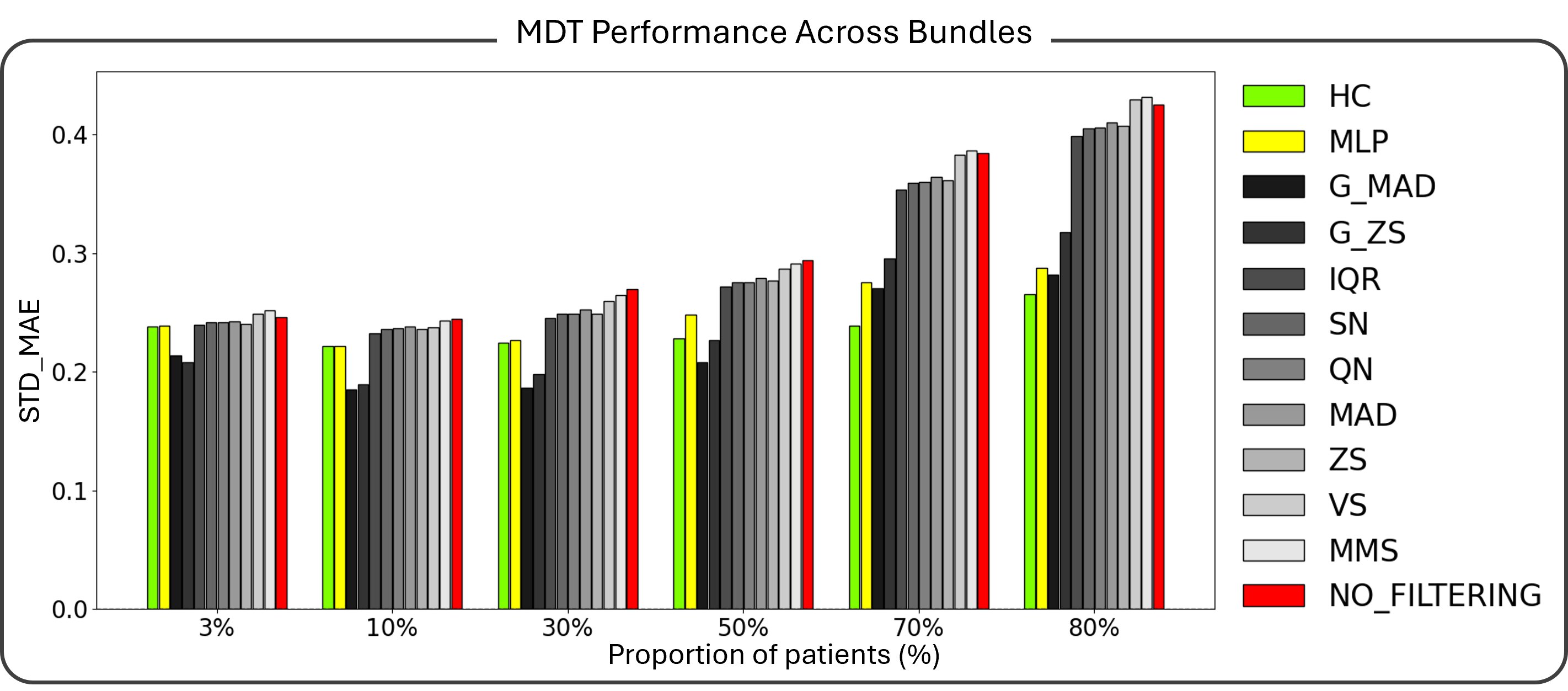} 
    \caption{This illustration is complementary to figure 3 in the paper.}
    \label{fig:S6}
\end{figure}

\begin{figure}[ht]
    \centering
    \includegraphics[width=0.99\textwidth]{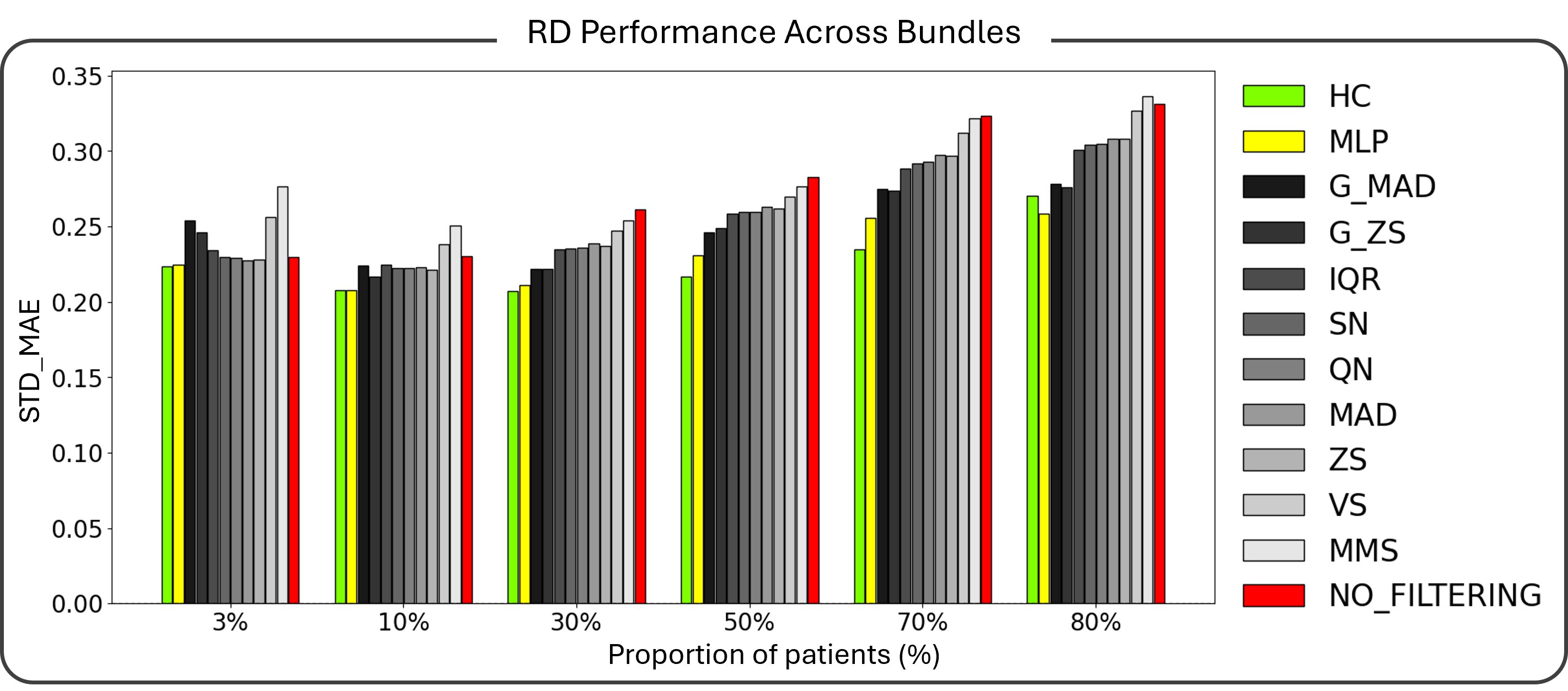} 
    \caption{This illustration is complementary to figure 3 in the paper.}
    \label{fig:S7}
\end{figure}

\begin{figure}[ht]
    \centering
    \includegraphics[width=0.99\textwidth]{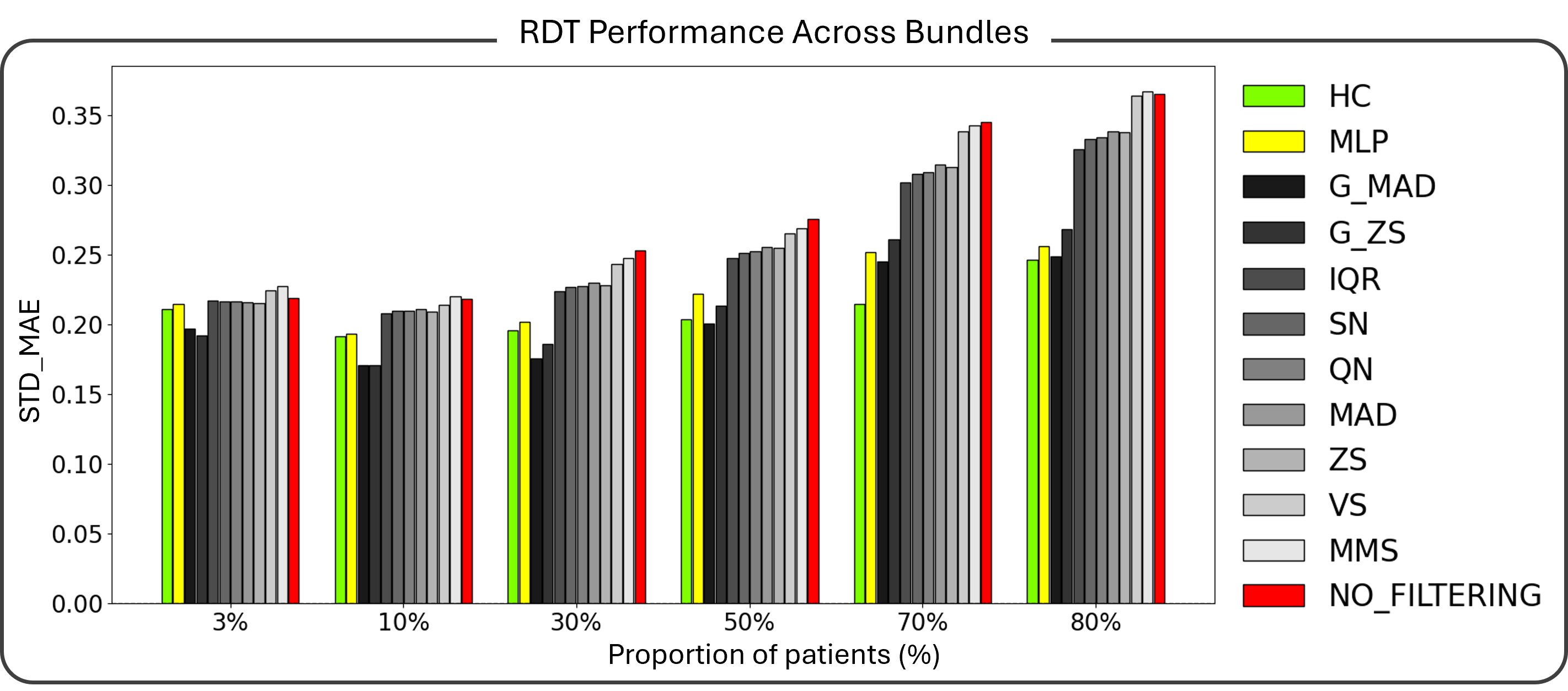} 
    \caption{This illustration is complementary to figure 3 in the paper.}
    \label{fig:S8}
\end{figure}